\documentclass[runningheads]{llncs}

\usepackage{graphicx}

\usepackage{tikz}
\usepackage{comment}
\usepackage{amsmath,amssymb} 
\usepackage{color}
\usepackage[accsupp]{axessibility}  

\usepackage{url}
\usepackage{wrapfig}
\usepackage{caption}
\usepackage{subcaption}
\usepackage[export]{adjustbox}
\usepackage{multirow}
\usepackage{multicol}
\usepackage{booktabs}
\usepackage{pifont}

\newcommand{\cmark}{\ding{51}}%
\newcommand{\xmark}{\ding{55}}%
\usepackage[capitalize]{cleveref}
\crefname{section}{Sec.}{Secs.}
\Crefname{section}{Section}{Sections}
\Crefname{table}{Table}{Tables}
\crefname{table}{Tab.}{Tabs.}

\newcommand{\modelname}{StARformer}
\newcommand{\repname}{StAR-representation}
\newcommand{\repnames}{StAR-representations}
\newcommand{\stepname}{Step Transformer}
\newcommand{\seqname}{Sequence Transformer}

\begin{document}
\pagestyle{headings}
\mainmatter
\def\ECCVSubNumber{4447}  

\title{StARformer: Transformer with State-Action-Reward Representations for Visual Reinforcement Learning} 


\titlerunning{StARformer}
%
\author{Jinghuan Shang\and
Kumara Kahatapitiya\and
Xiang Li\and
Michael S. Ryoo}
\authorrunning{J. Shang et al.}
\institute{Stony Brook University, NY 11794, USA \\
\email{\{jishang, kkahatapitiy, xiangli8, mryoo\}@cs.stonybrook.edu}\\
}
\maketitle

\begin{abstract}
Reinforcement Learning (RL) can be considered as a sequence modeling task: given a sequence of past state-action-reward experiences, an agent predicts a sequence of next actions. 
In this work, we propose \textbf{St}ate-\textbf{A}ction-\textbf{R}eward Transformer (\textbf{StAR}former) for visual RL, which explicitly models \textit{short-term} state-action-reward representations (StAR-representations), essentially introducing a Markovian-like inductive bias to improve \textit{long-term} modeling.
Our approach first extracts StAR-representations by self-attending image state patches, action, and reward tokens within a short temporal window. These are then combined with pure image state representations --- extracted as convolutional features, to perform self-attention over the whole sequence.
Our experiments show that StARformer outperforms the state-of-the-art Transformer-based
method on image-based Atari and DeepMind Control Suite benchmarks, in both offline-RL and imitation learning settings. StARformer is also more compliant with longer sequences of inputs.
Our code is available at \url{https://github.com/elicassion/StARformer}.
\end{abstract}

\section{Introduction}\label{sec:introduction}

Reinforcement Learning (RL) naturally operates sequentially: an agent observes a state from the environment, takes an action, observes the next state, and receives a reward from the environment. 
 In the past, RL problems have been usually modeled as Markov Decision Processes (MDP). It enables us to take an action solely based on the current state, which is assumed to represent the whole history.
 With this scheme, sequences are broken into single steps so that algorithms like TD-learning~\cite{td} can be mathematically derived via Bellman Equation to solve RL problems.
Recent advances such as~\cite{dt,trajectorytransformer} formulate (offline-)RL differently--- as a sequence modeling task, and Transformer~\cite{att} architectures have been adopted as generative trajectory models to solve it, i.e., given past experiences of an agent composed of a sequence of state-action-reward triplets, a model iteratively generates an output sequence of action predictions.

\begin{figure}[t]
    \centering
    \includegraphics[width=\linewidth]{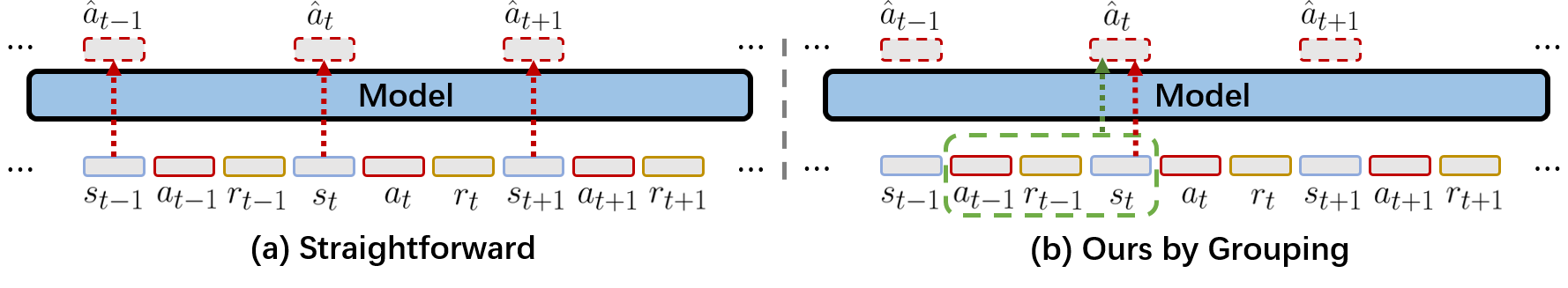}
    \caption{Illustration of RL as sequence modeling using Transformer: (a) A straightforward approach and, (b) Our proposed improvement. The intuition is to explicitly model local features (green boundary) to help long-term sequence modeling. 
    }
    \label{fig:seqmodelrl}
\end{figure}

This new scheme softens the MDP assumption, where an action is predicted considering multiple steps in history.
To implement this, methods such as~\cite{dt,trajectorytransformer} process the input sequence \textit{plainly} through self-attention (with a causal attention mask) using Transformers~\cite{att}.
This way, a given state, action, or reward token may attend to any of the (previous) tokens in the sequence, which allows the model to capture long-term relations. Moreover, each image state is usually encoded with convolutional networks (CNNs) \textit{as-a-whole} prior to self-attention.

\begin{figure}[t]
\begin{minipage}{.47\textwidth}
  \includegraphics[width=\linewidth,left]{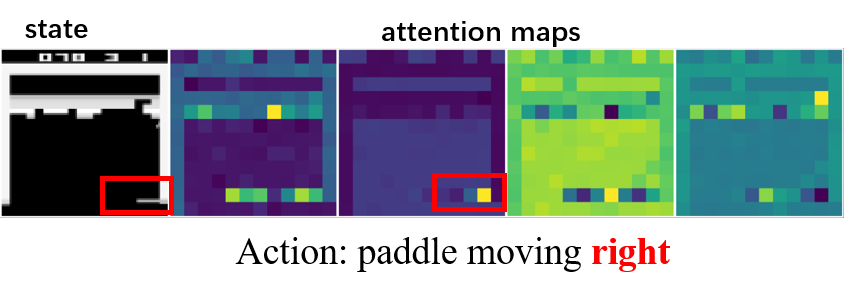}
  \captionof{figure}{Attention maps between action token and pixel state patches in our method. In the second attention map from the left, weights in the paddle region are directed towards right (highlighted in red), corresponding to the semantic meaning of ``right'' action.}
  \label{fig:vis_single}
\end{minipage}%
\hfill
\begin{minipage}{.47\textwidth}
  \includegraphics[width=\linewidth,right]{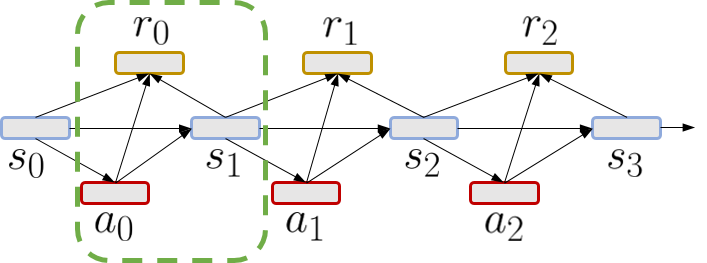}
  \captionof{figure}{MDP view of an RL process. Only the connected pairs (directed arrows) are causally-related, whereas others are independent of each other. 
    Green boundary highlights our motivation: explicitly considering a single transition helps long-term modeling.}
  \label{fig:mdp}
\end{minipage}
\end{figure}

However, if we consider states, actions, and rewards within adjacent time-steps, they generally have strong connections due to potential causal relations. For instance, states in the recent past have a stronger effect on the next action, compared to states in the distant past. Similarly, the immediate-future state and the corresponding reward are direct results of the current action.
In an extreme case--- MDP, the relations are far more strong and restricted (see Fig.~\ref{fig:mdp}).
In the above scenarios, a Transformer attending to all tokens naively may suffer from excess information (making the learning-process harder) or dilute the truly-essential relation priors. This is especially critical when input sequences are quite large, either in spatial~\cite{yang2021focal} or temporal~\cite{trajectorytransformer} dimension, and when Transformer models become heavy, i.e., contain a large number of layers~\cite{touvron2021going}.
Learning Markovian-like dependencies between tokens from scratch is hard and may waste computations~\cite{trajectorytransformer}, as the rest of the dependencies are possibly weaker.
Moreover, tokenizing image states as-a-whole based on CNNs further prohibits Transformer models from capturing detailed spatial relations.
Such loss of information can be critical especially in RL tasks with fine-grained regions-of-interest.

To alleviate such issues, we propose to explicitly model single-step transitions, introducing a Markovian-like inductive bias and relieving the capacity to be used for long sequence modeling. We introduce \textbf{St}ate-\textbf{A}ction-\textbf{R}eward Transformer (\textbf{StAR}former) for visual RL, 
which consists of two interleaving components: a \stepname~and a \seqname. The \stepname~learns local representations (i.e., \textbf{StAR}-representations) by self-attending state-action-reward tokens \textit{within a window of single time-step}. Here, image states are encoded as ViT-like~\cite{vit} patches, retaining fine-grained spatial information.
The \seqname~then combines \repnames~with pure image state representations (extracted as convolutional features) \textit{from the whole sequence} to make action predictions. Our experiments validate the benefits of \modelname~over prior work in both offline-RL and imitation learning settings, while also being more compliant with longer input sequences

Our contributions are as follows: we 
(1) propose to model single-step transitions in RL explicitly, relieving model capacity to better focus on long-term relations, (2) present a method to combine ViT-like image patches with action and reward token to retain fine-grained spatial information, and (3) introduce an architecture to fuse \repnames~over a long-sequence with our interleaving Step and Sequence Transformer layers. In particular, this allows modeling sequences of state-action feature representations at multiple different levels.

\section{Related Work}
\subsection{Reinforcement Learning to Sequence Modeling}
Reinforcement Learning (RL) is usually modeled as a Markov Decision Process (MDP). 
Based on this, single-step value-estimation methods have been derived from the Bellman equation, including Q-learning~\cite{qlearning} and Temporal Difference (TD) learning~\cite{sarsa,td,tdgammon,konda2000actor}, along with their extensions~\cite{dqn,sacae,hessel2018rainbow}. 

More recent directions~\cite{dt,trajectorytransformer} formulate RL a different way --- as a sequence modeling task, i.e., given a sequence of recent experiences including state-actions-reward triplets, a model predicts a sequence of next actions. This approach can be trained in a supervised learning manner, being more compliant with offline RL~\cite{levine2020offline} and imitation learning settings~\cite{gail,gailfo,shang2021self}.
Zheng et al.~\cite{zheng2022online} adapt this formulation to online settings.
Furuta et al.~\cite{furuta2022distributional} extend DT~\cite{dt} to match given hindsight information.
Reed et al.~\cite{reed2022generalist} train a single agent that performs a wide range of RL and language tasks.
Sequence modeling can be also viewed as solving RL by learning trajectory representations. Other than methods learning visual representations only~\cite{sacae,drq1,drq2,rad,laskin2020curl,li2022does,3dtrl}, our approach combines visual and trajectory representations together, thanks to the power of Transformer.

\subsection{Transformers}
Transformer architectures~\cite{att} have been first introduced in language processing tasks~\cite{bert,pretrainnlp,gpt2}, to model interactions between a sequence of word embeddings, or more generally, unit representations or \textit{tokens}. Recently, Transformers have been adopted in vision tasks with the key idea of breaking down images/videos into tokens~\cite{vit,arnab2021vivit,imagegpt,vtn,tnt,timesformer,kahatapitiya2021swat}, often outperforming convolutional networks (CNNs) in practice. Inspired by designs from both Transformers and CNNs, combining the two~\cite{dai2021coatnet,liu2022convnet} shows further improvements. Transformers also found to be useful in handling sensory information~\cite{tang2021sensory} and doing one-shot imitation learning~\cite{dasari2020transformers}. Chen et al.\cite{dt} explore how GPT~\cite{gpt2} can be applied to RL under the sequence modeling setting.

Sequence modeling in visual RL is similar to learning from videos in terms of input data, which are composed of sequences of observed images (i.e. states). One challenge of applying Transformers to videos is the large number of input tokens and quadratic computation. These problems have been investigated in multiple directions, including attention approximation~\cite{performer,linformer,kitaev2019reformer}, separable attention in different dimensions~\cite{arnab2021vivit,timesformer}, reducing the number of tokens using local windows~\cite{liu2021swin,videoswin}, adaptively generating a small amount of tokens~\cite{ryoo2021tokenlearner} or using a CNN-stem to come up with a small amount of high-level tokens \cite{xiao2021early,neimark2021video,dai2022ms}.

\modelname~shares a similar concept to performing spatial and temporal attention separately as in~\cite{arnab2021vivit,timesformer}.
In contrast to such methods designed to reduce attention computation, our primary target is introducing inductive bias: modeling short-term and long-term contexts separately.
Our method also operates on different sets of tokens, in short-term (s-a-r tokens) and in long-term (learned intermediate \repname), which deviates from previous methods.
\section{Preliminary}

\subsection{Transformer}\label{sec:pre-trans}
Transformer~\cite{att} architectures have shown diverse applications in language~\cite{bert} and vision tasks~\cite{vit,arnab2021vivit}.
Given a sequence of input tokens $X = \{x_1, x_2, ..., x_n\}$, where $x_i \in \mathbb{R}^d$, a Transformer layer maps it to an output sequence of tokens $Z = \{z_1, z_2, ..., z_n\}$, where $z_i \in \mathbb{R}^d$.
A Transformer model is obtained by stacking multiple such layers. We denote the mapping for each layer ($l$) as $F(\cdot)$: $Z^{l} = F(Z^{l-1})$. We use $F(\cdot)$ to represent a Transformer layer in the remaining sections.

Self-attention~\cite{lin2017structured,parikh2016decomposable,cheng2016long,att} is the core component of Transformers, which models pairwise relations between tokens. As introduced in~\cite{att}, an input token representation $X$ is linearly mapped into query, key and value representations, i.e., $\{Q, K, V\} \in \mathbb{R}^{n\times d}$ respectively, to compute self-attention as follows:
\begin{equation}
    \text{Attention}(Q,K,V) = \text{softmax}(\frac{QK^T}{\sqrt{d}})V.
\end{equation}

Vision Transformer (ViT)~\cite{vit} extends the same idea of self-attention to the image domain.
Given an input image $s \in \mathbb{R}^{H\times W\times C}$, a set of $n$ non-overlapping local patches $P = \{p_i\} \in \mathbb{R}^{h\times w\times C}$ is extracted, flattened and linearly mapped to a sequence of tokens $\{x_i\} \in \mathbb{R}^{d}$.
We extend ViT~\cite{vit} so that action, reward, and state patches can jointly attend, where we find semantic meanings could be learned within action-patch attention in RL tasks (Fig.~\ref{fig:seqmodelrl}).

\subsection{RL as Sequence Modeling}
We consider a Markov Decision Process (MDP), described by tuple $(\mathcal{S}, \mathcal{A}, P, \mathcal{R})$, where $s\in \mathcal{S}$ represents the state, $a\in \mathcal{A}$, the action, $r\in \mathcal{R}$, the reward, and $P$, the transition dynamics given by $P(s'|s, a)$.
In MDP, a trajectory ($\tau$) is defined as the past experience of an agent, which is a sequence composed of states, actions, and rewards in the following temporal order:
\begin{equation}
    \tau = \{ s_1,\; a_1,\; r_1,\;  s_2,\; a_2,\; r_2,\; \dots ,\;  s_t,\; a_t,\; r_t \}.
\end{equation}
Sequence modeling for RL is making action predictions from past experience~\cite{dt,trajectorytransformer}:
\begin{equation}
    Pr(\hat{a}_t) = p(a_t|\; s_{1:t},\; a_{1:t-1}, \;r_{1:t-1}).
    \label{eq:causal}
\end{equation}

Recent work~\cite{dt,trajectorytransformer} try to adopt an existing Transformer architecture~\cite{gpt2} for RL with the formulation as above.
In~\cite{dt,trajectorytransformer}, states ($s$), actions ($a$), and rewards ($r$) are considered as input tokens (see Fig.~\ref{fig:seqmodelrl}a),
while using a causal mask to ensure an autoregressive output sequence generation (i.e. following Eq.~\ref{eq:causal}), where a token can access any of its preceding tokens through self-attention.

In contrast, our formulation attends tokens with (potentially) strong causal relations \textit{explicitly}, while attending to long-term relations as well.
To do this, in this work, we break a trajectory into small groups of state-action-reward tuples (i.e., $s, a, r$). It learns local relations within the tokens of each group through self-attention (see Fig.~\ref{fig:seqmodelrl}b, and Fig.~\ref{fig:vis_single}), followed by long-term sequence modeling. 

\section{StARformer}
\begin{figure}[t]
    \centering
    \includegraphics[width=\textwidth]{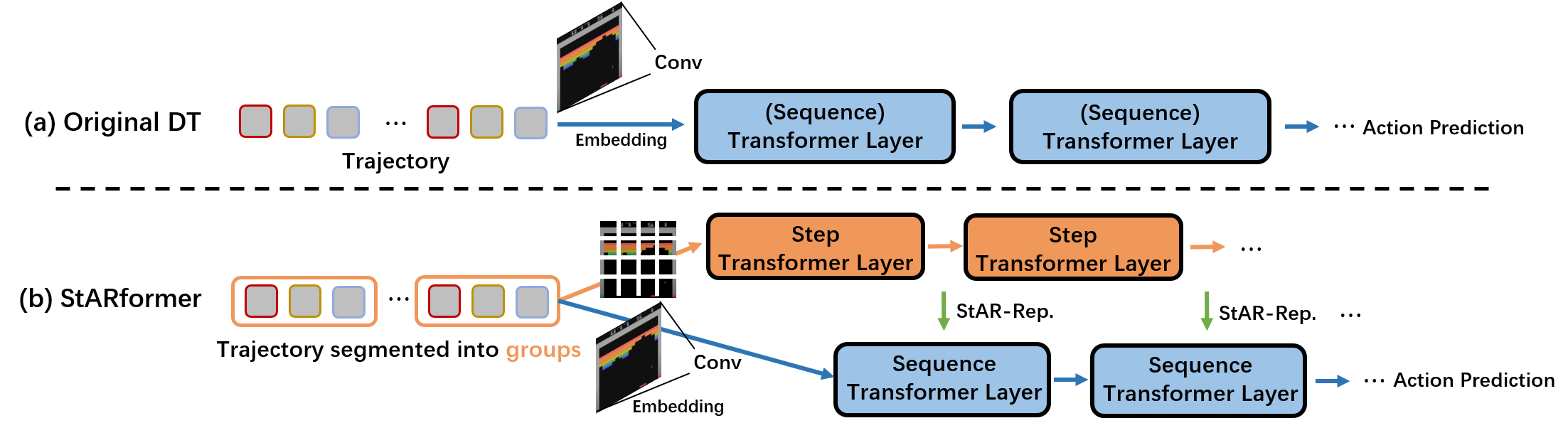}
        \caption{(a) Structure summary of original DT~\cite{dt}, where the Transformer Layer acts similar as our \seqname. (b) \modelname~consists of \stepname~and \seqname, to separately model a single-step and the sequence as-a-whole, respectively. Two types of layers are connected at each level via learned \repnames. In terms of state embedding methods, DT uses only convolution, while \modelname~uses ViT-like~\cite{vit} embeddings (patches) in \stepname~and convolution in \seqname~separately.}
    \label{fig:main_fig}
\end{figure}
\subsection{Overview}
\modelname~consists of two basic components: \stepname~and~\seqname, together with interleaving connections (see Fig.~\ref{fig:main_fig}b). \stepname~learns \repname{s}~from strongly-connected local tokens \textit{explicitly}, which are then fed into the \seqname~along with pure state representations to model the whole input trajectory.
At the output of the final \seqname~layer, we make action predictions via a prediction head.
In the following subsections, we will introduce the two Transformer components, and their corresponding token embeddings in detail.

\subsection{\stepname}
\paragraph{Grouping State-Action-Reward:} 
\begin{wrapfigure}[21]{r}{0.3\textwidth}
    \centering
    \includegraphics[width=0.24\textwidth]{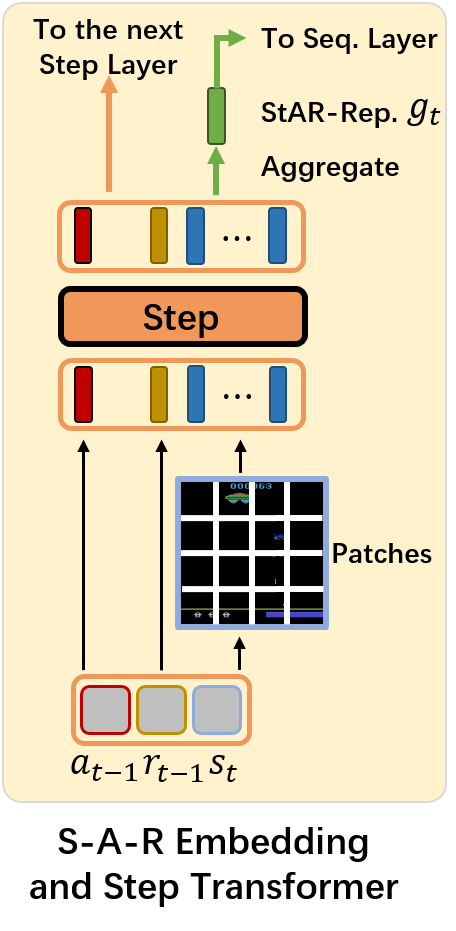}
    \caption{Overview of \stepname. Output tokens are (1) sent to the next \stepname~layer and (2) aggregated to produce \repname. 
    }
    \label{fig:sarandstep}
\end{wrapfigure}
Our intuition of grouping is to model strong local relations explicitly. To do so, we first segment a trajectory ($\tau$) into a set of groups, where each group consists of previous action ($a_{t-1}$), reward ($r_{t-1}$) and current state ($s_t$) (Fig.~\ref{fig:sarandstep}). Each element within a group has a strong causal relation with the others.

\paragraph{Patch-wise State Token Embeddings:} In \stepname, we tokenize each input image state by dividing it into a set of non-overlapping spatial patches $z_{s_t}$ along its spatial dimensions, following ViT~\cite{vit} (Fig.~\ref{fig:sarandstep}).
Our motivation for using patch embeddings is to create fine-grained state embeddings.
This allows the \stepname~to model the relations of actions and rewards with local-regions of state (Fig~\ref{fig:vis_single}).
Such local correspondences provide more information compared to highly-abstracted convolutional features in this single-step modeling, which is empirically validated in our ablation studies (Sec.~\ref{sec:patch}).

\paragraph{Action and Reward Token Embeddings:}
We embed the action and reward tokens with a linear layer as in~\cite{dt}.
\paragraph{S-A-R embeddings:}
Altogether, we get a collection of state, action, and reward embeddings as the input to the initial \stepname~layer which is given by:
    $Z^0_t = \{ z_{a_{t-1}},\; z_{r_t},\; z_{s^1_t},\; z_{s^2_t},\; \dots,\; z_{s^n_t} \}.$
We have $T$ groups of such token representations per trajectory, which are simultaneously processed by the \stepname~with shared parameters. 

\paragraph{\stepname~Layer:}
We adopt the conventional Transformer design from ~\cite{att} (Sec.~\ref{sec:pre-trans}) as our \stepname~layer.
Each group of tokens from the previous layer $Z^{l-1}_t$ is transformed to $Z^l_t$ by a \stepname~layer with the mapping $F^{l}_{\text{step}}$: 
$Z^l_t = F^l_{\text{step}}(Z^{l-1}_t)$.
\paragraph{\repname:}
At the output of each \stepname~layer $l$, we further obtain a State-Action-Reward-representation (\repname) $g^l_t \in \mathbb{R}^D$ by aggregating output tokens $Z^l_t \in \mathbb{R}^{n\times d}$ (see green flows in Fig.~\ref{fig:main_fig}b):
\begin{equation}
    g^l_t = \text{FC}([Z^l_t]) + e^{\text{temporal}}_t.
\end{equation}
Here $[\cdot]$ represents concatenation of the tokens within each group and $e^{\text{temporal}}_t \in \mathbb{R}^{D}$, the temporal positional embeddings for each timestep.
Finally, the output \repname~$g^l_t$ is fed into the corresponding \seqname~layer for long-term sequence modeling.

\subsection{\seqname}

\begin{figure}[t]
    \centering
    \includegraphics[width=0.98\textwidth]{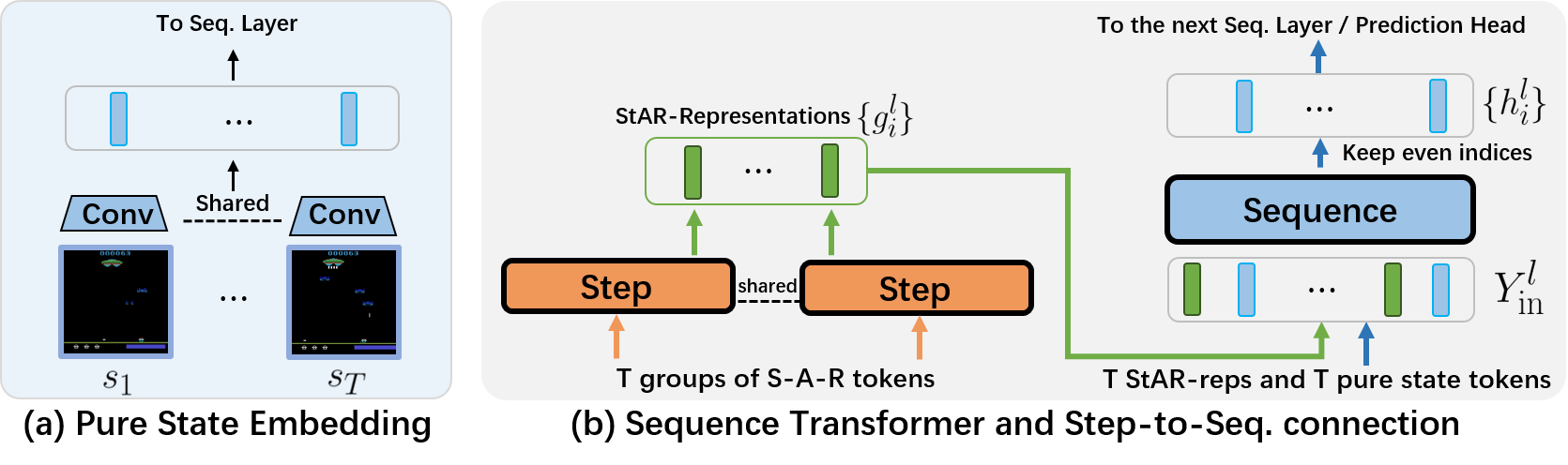}
    \caption{(a) Pure state embeddings are learned from shared convolutional layers. (b) \seqname~takes \repname~and the pure state tokens and generate output tokens. 
    }
    \label{fig:pseandseq}
\end{figure}

Our \seqname~models long-term sequences by looking at the learned~\\ \repnames~and the \textit{pure state tokens} (introduced below) over the whole trajectory (See Fig.~\ref{fig:pseandseq}). Notice that, as illustrated in Fig.~\ref{fig:main_fig} (b), this happens with multiple intermediate StAR representations, allowing the \seqname~to capture detailed information. 

\paragraph{Pure State Token Embeddings:} 
In addition to the patch-wise token embeddings in \stepname, we embed the input image state $s_t$ as-a-whole, to create pure state tokens $h^0_t$. Each such token represents a single state representation, describing the state globally in space.
We do this by processing each state through a CNN encoder, since the convolutional layers mix features spatially:
\begin{equation}
    h^0_t = \text{Conv}(s_t) + e^{\text{temporal}}_t ,
\end{equation}
where $e^{\text{temporal}}_t \in \mathbb{R}^{D}$ represents the temporal positional embeddings exactly the same as we add to $g_t$ for each timestep.
The convolutional encoder is from~\cite{natureencoder}.

\paragraph{\seqname~Layer:}
Similar to \stepname, we use the conventional Transformer layer design from~\cite{att} for our \seqname.
The input to the \seqname~layer $l$ consists of representations from two sources: (1) the learned \repname{s}~$g^l_t\in \mathbb{R}^{D}$ from the corresponding \stepname~layer, and (2) $h^{l-1}_t \in \mathbb{R}^D$ from the outputs of the previous \seqname~layer. Here, as mentioned above, we set $h^0_t$ to be the pure state representation.
The two types of token representations are merged to form a single sequence, preserving their temporal order (as elaborated below):
\begin{equation}\label{eq:y_in}
    Y_{\text{in}}^{l} = \{g^l_1,\; h^{l-1}_1,\; g^l_2,\; h^{l-1}_2,\; \dots,\; g^l_T,\; h^{l-1}_T\}.
\end{equation}
We place $g^l_t$ before $h^{l-1}_t$--- which originates from $s_t$--- because $g^l_t$ contains information of the \textit{previous} action $a_{t-1}$, which comes prior to $s_t$ in the trajectory. We also apply a causal mask in the \seqname~to ensure that the tokens at time $t$ cannot attend any future tokens (i.e., $>t$).

Here, \seqname~takes \repnames~generated from each intermediate \stepname~layer, rather than taking the final StAR-represent-ations after all \stepname~layers. In this way, the model gains an ability to look at \repnames~at multiple abstraction levels. 
In Section~\ref{sec:connection}, we empirically validate the benefit of this layer-wise fusion.

\seqname~computes an intermediate set of output tokens as in: $Y_{\text{out}}^{l} = F^l_{\text{sequence}}(Y_{\text{in}}^{l}).$
We then select the tokens at even indices of $Y_{\text{out}}^{l}$ (where indexing starts from 1) to be the pure state tokens $h^{l}_i := {y_{\text{out};2i}^{l}}$, which are then fed into the next \seqname~layer.


\paragraph{Action Prediction:}
The output of the last \seqname~layer is used to make action predictions, based on a linear head: $\hat{a}_t = \phi(h^l_{t})$.

\subsection{Training and Inference}\label{sec:training}
\modelname~is a drop-in replacement of DT~\cite{dt}, as training and inference procedures remain the same.
\modelname~can easily operate on step-wise reward without a performance drop (detailed discussed in \ref{sec:rewardsetting}). In contrast, it is critical to design a Return-to-go (RTG, target return) carefully in DT, which needs more trials and tuning to find the best value.


\section{Experimental Setup}\label{sec:exp}
\begin{figure*}[t]
    \centering
    \includegraphics[width=\linewidth]{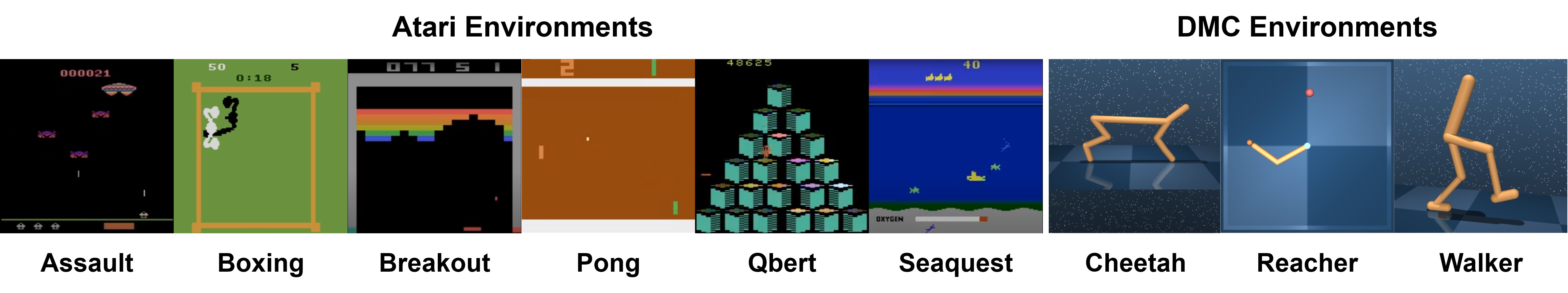}
    \caption{Environments used: Atari is with a discrete action space, and DMC is with a continuous action space. We use gray-scale input similar to prior work~\cite{dqn,dt}.}
    \label{fig:simulationenvironment}
\end{figure*}

\subsection{Settings}
We consider offline RL~\cite{levine2020offline} and imitation learning (behavior cloning) in our experiments.
In offline RL, we have a fixed memory buffer of sub-optimal trajectory rollouts.
Offline RL is generally more challenging compared to conventional RL due to the shifted distribution problem~\cite{levine2020offline}.

In our Imitation learning setting, the agent is not exposed to reward signals and online-collected data from the environment. 
This is an even harder problem due to provided trajectories being sub-optimal, compared to traditional imitation learning that could collect new data and do Inverse Reinforcement Learning~\cite{ng2000algorithms,abbeel2004apprenticeship}.
We simply remove the rewards in the dataset used in offline RL to come up with this setting.
Both DT~\cite{dt} and proposed method can operate without reward, by simply removing  $T$ reward tokens in DT~\cite{dt} ($T$ is trajectory length), or removing the reward token in \stepname~in our model.

\subsection{Environments and Datasets}
We consider image-based Atari~\cite{atari} (discrete action space) and DeepMind Control Suite (DMC)~\cite{tassa2020dmcontrol} (continuous action space) to evaluate our model in different types of tasks, listed in Fig.~\ref{fig:simulationenvironment} with image examples.
We pick 6 games in Atari: Assault, Boxing, Breakout, Pong, Qbert, and Seaquest.
Similar to~\cite{dt} we use 1\% (500k steps) of the DQN replay buffer dataset~\cite{dqndataset} to perform a thorough and fair comparison.
We select 3 continuous control tasks in DMC~\cite{tassa2020dmcontrol}: Cheetah-run, Reacher-easy, and Walker-walk.
In DMC, we collect a replay buffer (i.e. sub-optimal trajectories) generated by training a SAC~\cite{haarnoja2018soft} agent from scratch for 500k steps for each task.
Note that these continuous control tasks are with image inputs, which previous work~\cite{dt} does not cover (originally using Gym~\cite{openaigym}).

We report the absolute value of episodic returns (i.e., cumulative rewards).
Results are averaged across 7 random seeds in Atari and 10 seeds in DMC, each seed is evaluated by 10 randomly initialized episodes.

\subsection{Baselines}
We select Decision-Transformer (DT)~\cite{dt}, a SOTA Transformer-based sequence modeling method for RL. We notice there is also Trajectory-Transformer~\\
\cite{trajectorytransformer}, which however, is not designed for image inputs. We use most of the same hyper-parameters as in DT~\cite{dt} for Atari environments without extra tuning (details in Supplementary Table 4 and 5).
As for DMC environments, since they are not covered by DT~\cite{dt}, we carefully tune the baseline first and then use the same set of hyper-parameters in our method.
We also compare with SOTA non-Transformer offline-RL methods including CQL~\cite{cql}, QR-DQN~\cite{qrdqn}, REM~\cite{rem}, and BEAR~\cite{bear}. For imitation (behavior cloning), we only compare with DT~\cite{dt} and straightforward behaviour cloning with ViT (referred to as BC-ViT). 
\section{Results}\label{sec:res}
\begin{figure}[t]
    \centering
    \includegraphics[width=0.48\linewidth]{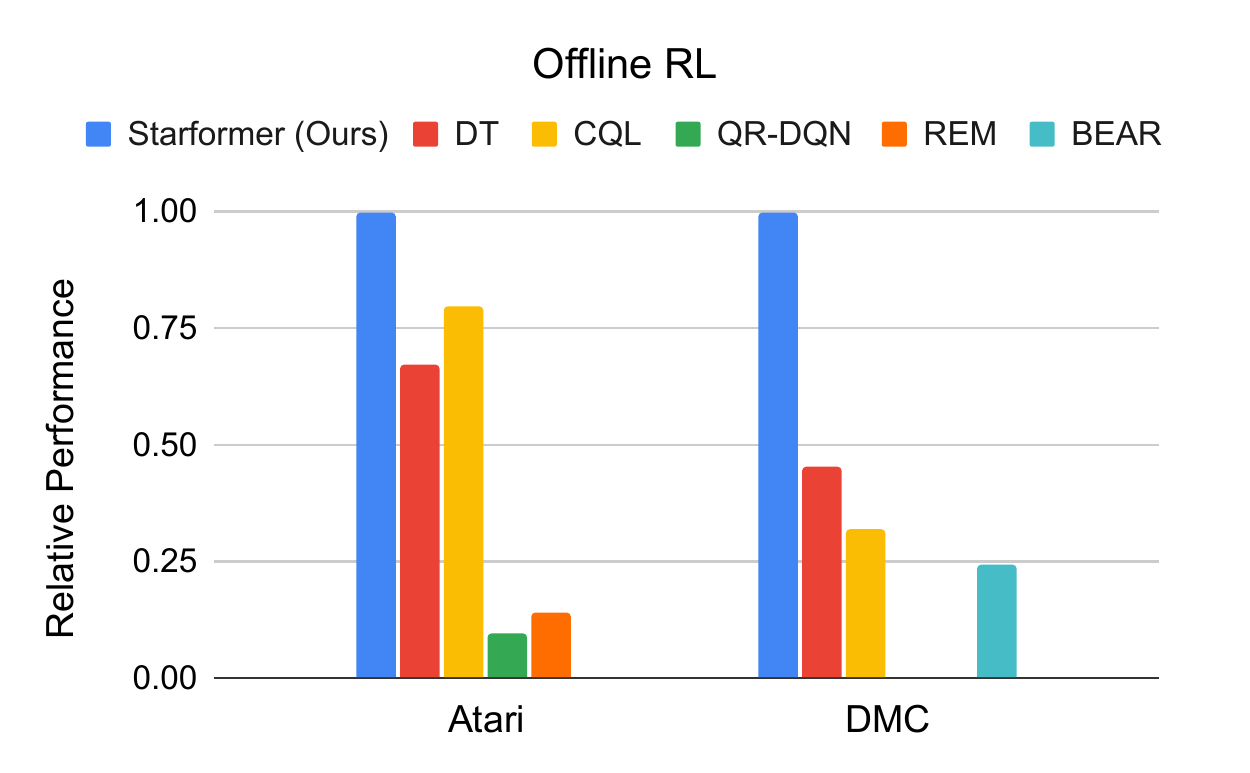}
    \includegraphics[width=0.48\linewidth]{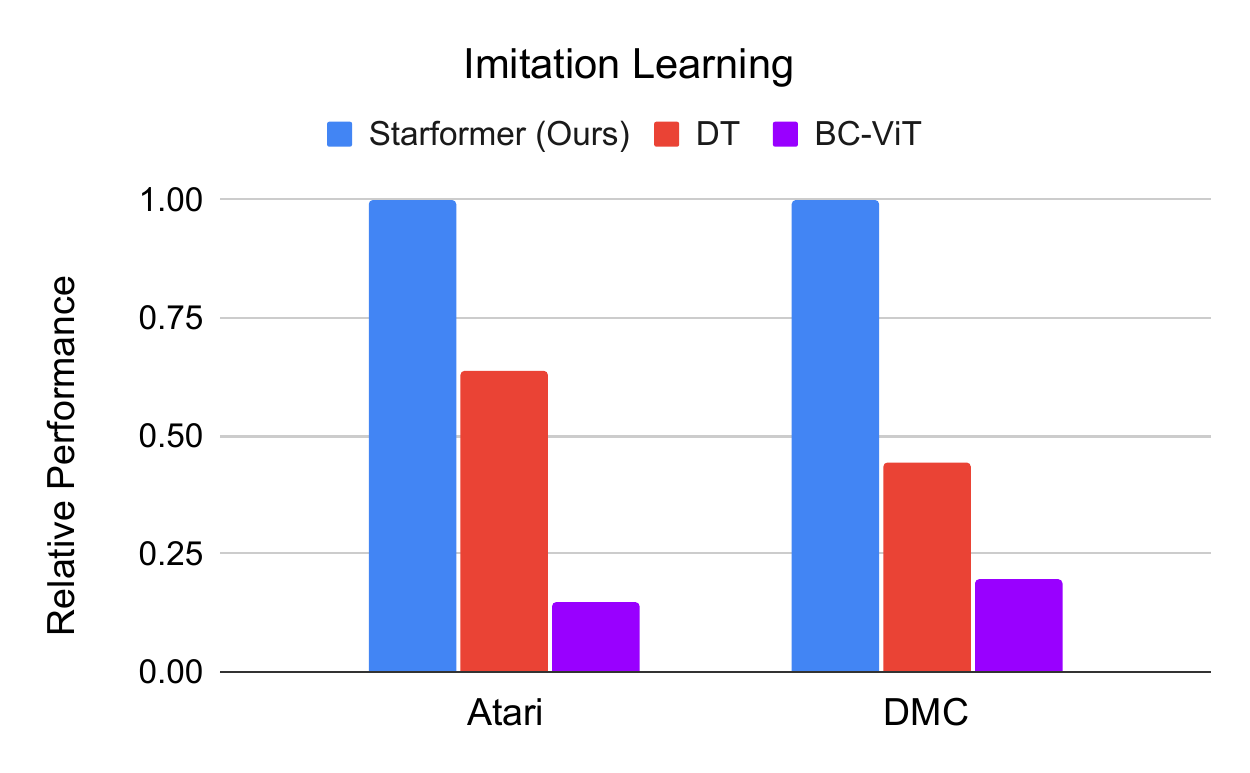}
    \caption{Relative performance of episodic returns. 
    The results are averaged across all environments and random seeds (same in later experiments), and normalized w.r.t.~the performance of StAR. 
    Please refer Table 1 in supplementary for details.}
    \label{fig:mainres_sum}
\end{figure}

\subsection{Improving Sequence Modeling for RL}\label{sec:mainres}
We first compare our \modelname~(StAR) with the state-of-the-art Transformer-based RL method in Atari and image-based DMC environments, under both offline RL and imitation learning settings.
We select the Decision-Transformer proposed in~\cite{dt}, (referred as DT) as our baseline. Here, we keep $T=30$ for all environments, which is the number of time-steps (length) of each input trajectory. 
We also compare our method to CQL~\cite{cql}, a SOTA non-Transformer offline-RL method.
Fig.~\ref{fig:mainres_sum} shows that our method outperforms baselines, in both offline RL and imitation learning settings, suggesting that our method can better model reinforcement sequences with images.



\subsection{Scaling-up to Longer Sequences}\label{sec:traj}
\begin{figure}[t]
    \centering
    \includegraphics[width=0.48\linewidth]{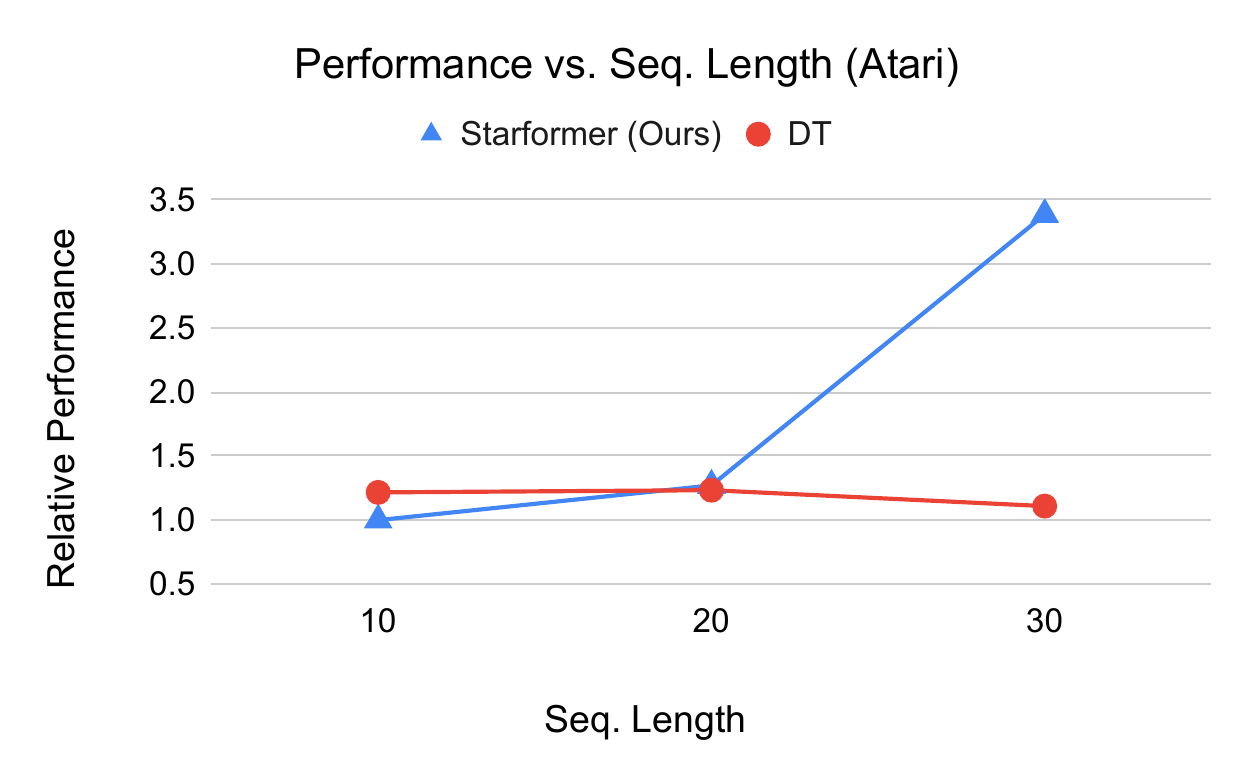}
    \includegraphics[width=0.48\linewidth]{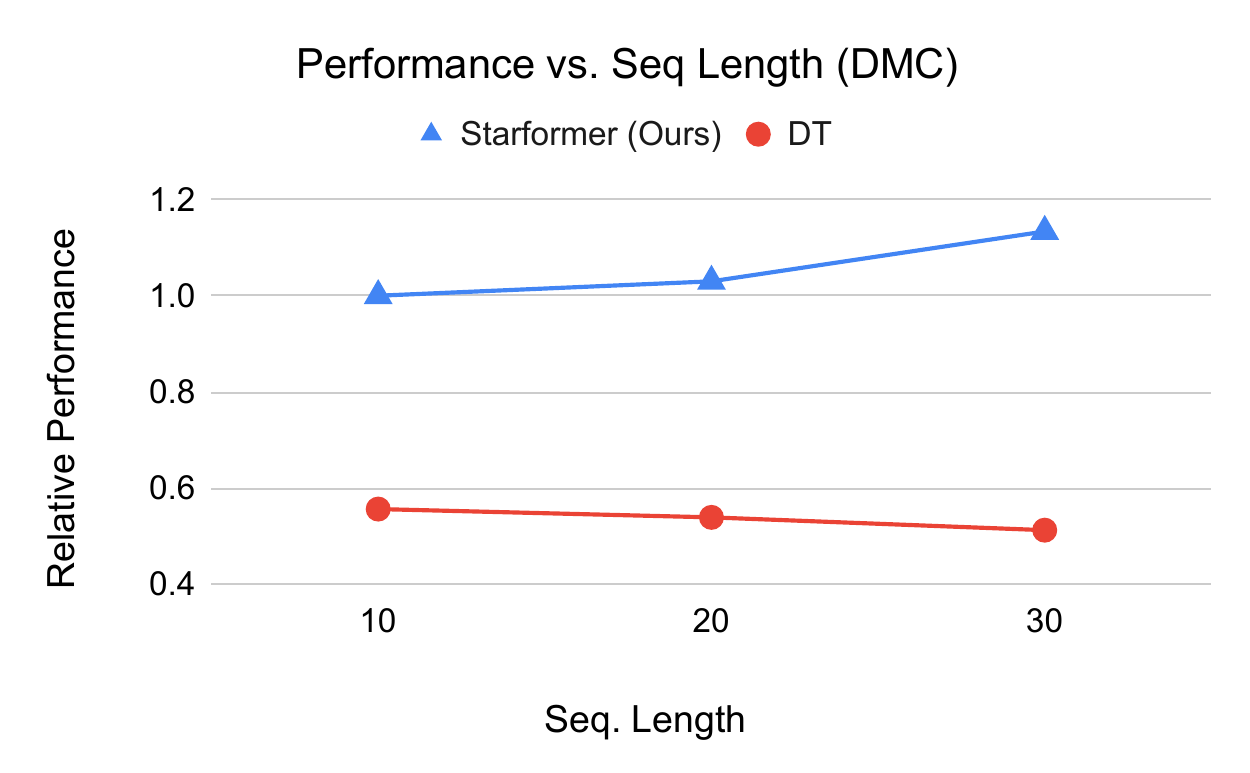}
    \caption{Change in performance with the length of input sequence, $T \in \{10, 20, 30\}$, in Atari and DMC (averaged across tasks), under offline-RL. 
    Please refer to Fig.1 in supplementary for per-task result.}
    \label{fig:seqlength_sum}
\end{figure}
In this experiment, we evaluate how \modelname~and DT perform with different input sequence lengths, specifically $T=\{10, 20, 30\}$, under offline-RL setting.
In Fig.~\ref{fig:seqlength_sum}, we see that \modelname~gains performance with longer trajectories, whereas DT~\cite{dt} saturates as early as $T=10$.
This validates our claim that considering short-term and long-term relations separately (and then fusing) help models to scale-up to longer sequences. Instead of learning Markovian pattern attentions~\cite{trajectorytransformer} implicitly, we model it explicitly in our \stepname. This acts as an inductive bias, relieving the capacity of \seqname~to better focus on long-term relations.
In contrast, DT takes off-the-shelf language model GPT~\cite{gpt2}, in which Markov property is not considered. 
\subsection{Visualization}
\begin{figure}[t]
    \centering
    \includegraphics[width=0.8\linewidth]{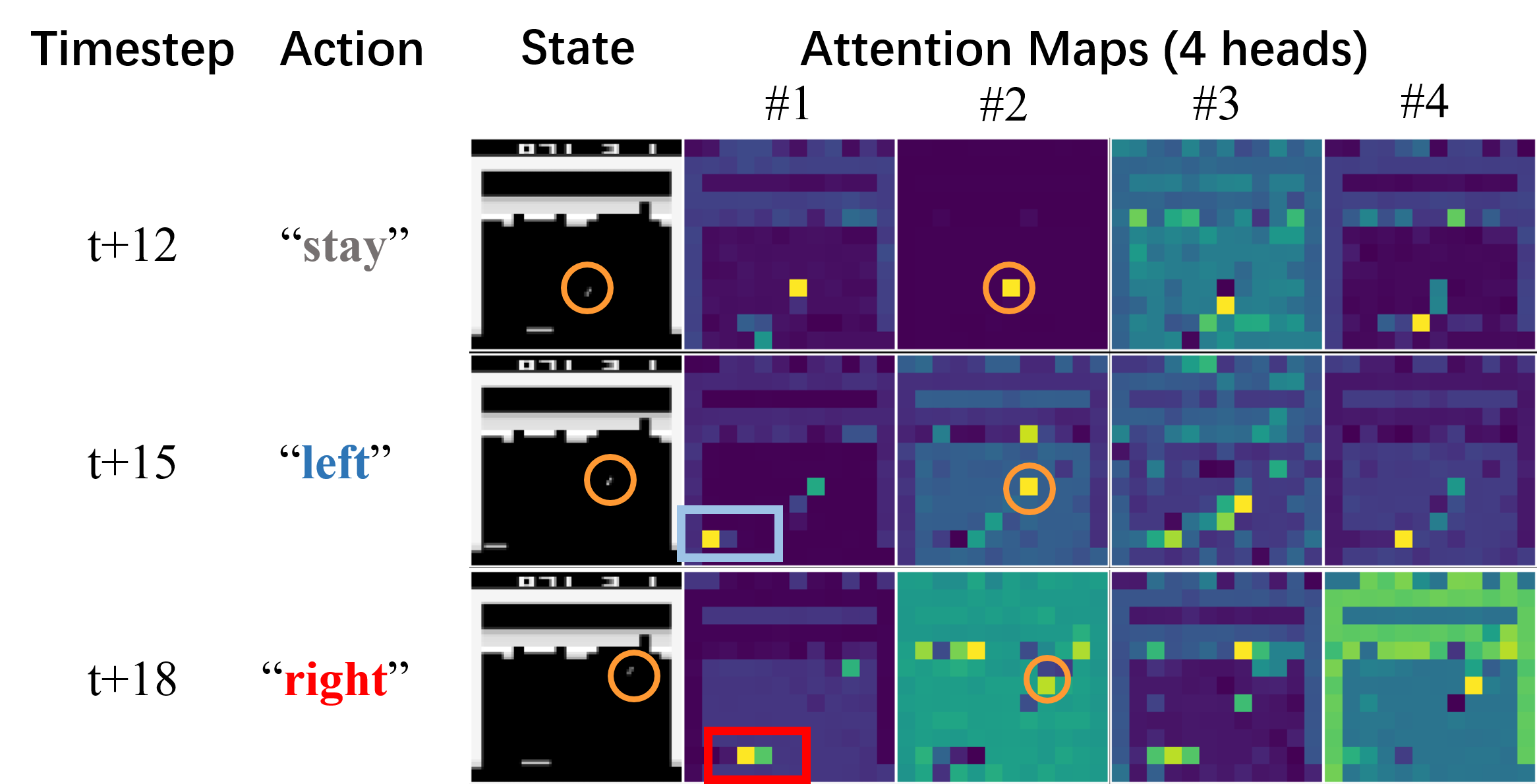}
    \caption{Visualization of attention maps in our \stepname, extracted for Breakout game.
    Attention weights are computed between the action token and state patch tokens. 
    We highlight the ball (orange circle) in the input for convenience. 
    Please find out more visualizations in our supplementary material. }
    \label{fig:vis}
\end{figure}

We show attention maps between action and state patches in \stepname~at several timesteps extracted from a trajectory in Breakout (see Fig.~\ref{fig:vis}).
In this game, the agent should move the paddle to bounce the ball back from the bottom, after the ball falls down while breaking the bricks on the top.
In the presented attention maps, the regions with a high attention score  (highlighted) mainly overlap with the locations of the ball, paddle, and potential target bricks.
We find the attention maps in head \#1 to be particularly interesting. Here, the focused regions corresponding to the paddle show a directional pattern, corresponding to the semantic meaning of actions ``moving the paddle right'', ``left'' or ``stay''. 
This validates that \stepname~captures essential spatial relations between actions and state patches, which is important for decision making. Moreover, in head \#2, we observe that the focused regions correspond to the locations of the ball, except when the ball is out of the boundary, too-close to the paddle or indistinguishable within bricks.
Overall, these attention maps suggest how our model can show a basic understanding of the Breakout game.

\begin{figure}[t]
    \centering
    \includegraphics[width=1.0\linewidth]{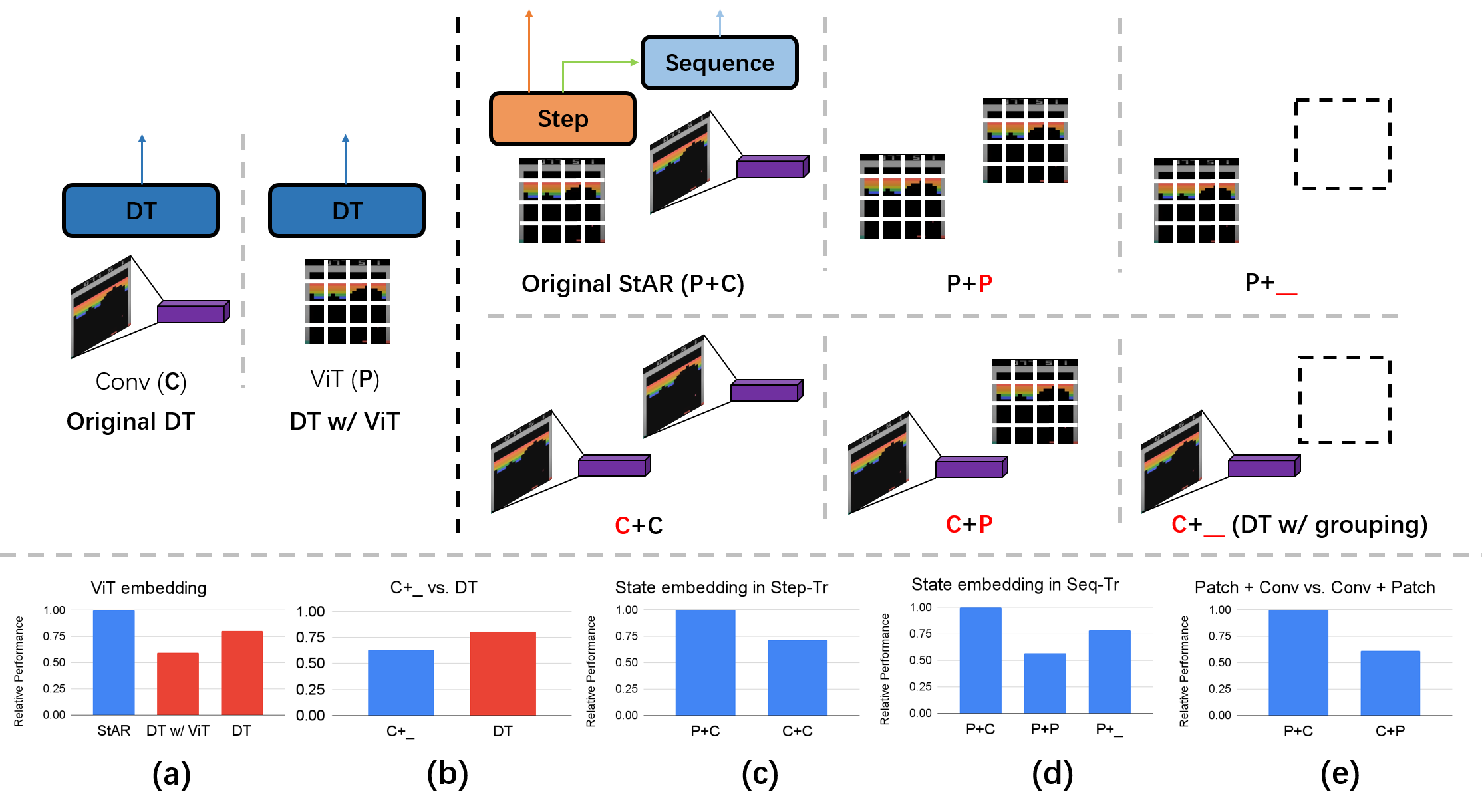}
    \caption{(Top): Embedding methods used in original DT, \modelname~(StAR), and their variants. We label ViT (patches) as \textbf{P}, convolution as \textbf{C}, and None (not using the corresponding embedding) as ``\textbf{\_\_}''. (Bottom): (a-e) Performance comparisons between variants. 
    Per-task results are in Supplementary Table 2.}
    \label{fig:ablation}
\end{figure}

\subsection{Ablations}

\modelname~has three design differences compared to baseline DT~\cite{dt}: it (1) learns \repname~from single-step transitions (grouping), (2) uses both ViT-like~\cite{vit} patch embeddings and convolutions for state representation, and (3) merges these two types of embeddings in \stepname~\textit{layer-wise}.
\subsubsection{\repname~and State Representations:}\label{sec:patch}
We first discuss designs of \repname~and state representation methods ((1) and (2) mentioned above) jointly, as they can be unified into variants shown in Fig.~\ref{fig:ablation}.
We vary state embedding methods used to learn $s_t$ in \stepname~and $h_t$ in \seqname. Namely, we consider: (1) ViT features (patch embeddings, labeled as \textbf{P}), (2) Convolutional features (labeled as \textbf{C}), or (3) None (not having the corresponding embedding, labeled as ``\textbf{\_\_}'' ).
The original \modelname~can be represented by \textbf{P}+\textbf{C} (patch embeddings for $s_t$ and convolutional embeddings for $h_t$ ). Other variants include: \textbf{P}+\textbf{P},$\;$ \textbf{P}+\textbf{\_\_},$\;$ \textbf{C}+\textbf{P},$\;$ \textbf{C}+\textbf{C},$\;$ and \textbf{C}+\textbf{\_\_}.
We note variant \textbf{C}+\textbf{\_\_} could be viewed as DT + grouping, where we simply adapt DT to our framework, and learn \repname~from convolutional features only.
We also implement a variant of DT using ViT for state embedding (noted as DT w/ ViT), to match our method in terms of having a similar embedding method and capacity (13M parameters vs. 14M parameters in ours).

When comparing \modelname~with original DT and DT w/ ViT (Fig.~\ref{fig:ablation}(a)), we see a performance drop in DT when used with ViT, which suggests that replacing convolutional features with ViT-like features naively would not benefit the model, despite the increased capacity (similar to ours). \modelname, however, does not benefit only from the larger capacity, but also from its better structural design, as verified in following experiments (see Fig.~\ref{fig:ablation}(c)(d)(e)). From Fig.~\ref{fig:ablation}(b), we see that \textbf{C}+\textbf{\_\_} which only uses convolutional features at \stepname, performs worse compared to DT. This is because convolutional features are highly abstracted, which makes them not well-suited for single-step transition (i.e., fine-grained) modeling.

When comparing \textbf{P}+\textbf{C} with \textbf{C}+\textbf{C} (Fig.~\ref{fig:ablation}(c)), the lower performance of \textbf{C}+\textbf{C} suggests that patches embeddings are better suited to model single transitions in \stepname.
In Fig.~\ref{fig:ablation}(d), we compare \textbf{P}+\textbf{C} with \textbf{P}+\textbf{P} and \textbf{P}+\textbf{\_\_}. We find convolution features work best in \seqname, validating that they provide abstract global information which is useful for long range modeling (coarse), in contrast to patch embeddings.
The observations from above comparisons of \modelname~variants suggest that our method benefits from fusing patch and convolutional features. 
We further evaluate this by comparing \textbf{P}+\textbf{C} and \textbf{C}+\textbf{P} (Fig.~\ref{fig:ablation}(e)), where \textbf{P}+\textbf{C} performs better, confirming this fusion method of ``fine-grained (patches) to high-level (conv)'' best matches with our sequence modeling scheme of ``single-transition followed-by long-range-context''.

\subsubsection{Step-to-Sequence Layer-wise Connections:}\label{sec:connection}
\begin{figure}[t]
    \centering
    \includegraphics[width=\textwidth]{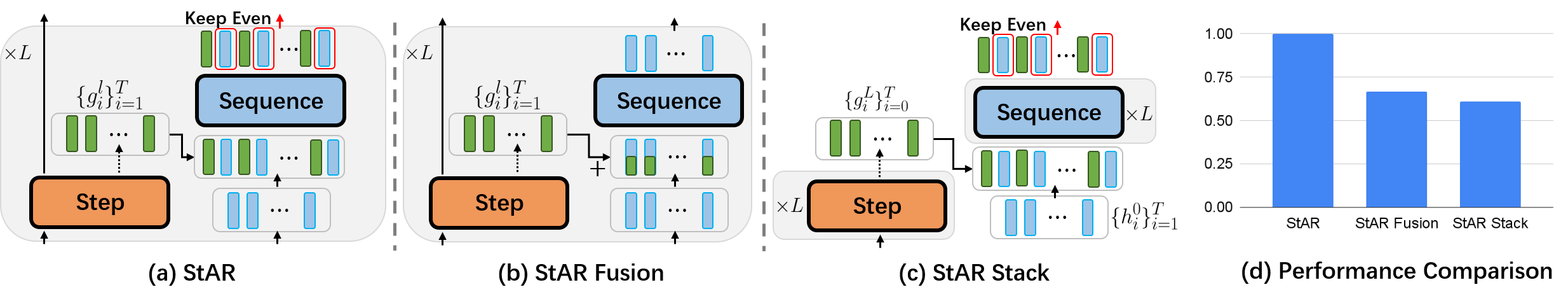}
    \caption{Variants of our model w/ different connections. Two variants, (b) StAR Fusion and (c) StAR Stack are shown in comparison to our original (a) StAR model. 
    (d) Experiments (offline-RL) show original structure, which is a layer-wise fusion, works best. 
    Please refer to Supplementary Table 3 for per-task results.
    }
    \label{fig:structures}
\end{figure}
In our model, we model whole trajectory using representations from two sources: \repnames~$g$ from \stepname~, and pure state representation $h$ from previous layer of \seqname.
We combine $g$ and $h$ in a layer-wise manner (i.e., at each corresponding layer). We investigate two other variants: (1) $g^l_t$ is fused with $h^l_t$ by summation (referred as StAR Fusion, see Fig.~\ref{fig:structures}(b)), and
(2) the \seqname~is ``stacked'' on-top of the \stepname~(referred as StAR Stack, see Fig.~\ref{fig:structures}(c)). 
Results of these configurations are shown in Fig.~\ref{fig:structures}(d) and StAR works the best. We see that attending to all tokens is better than token summation at \seqname. Also, having \repnames~from different abstraction levels is beneficial compared to having one.

\subsubsection{Reward setting: Return-to-go, stepwise reward, or no reward at-all?}\label{sec:rewardsetting}


\begin{figure}[t]
    \centering
    \includegraphics[width=0.7\linewidth]{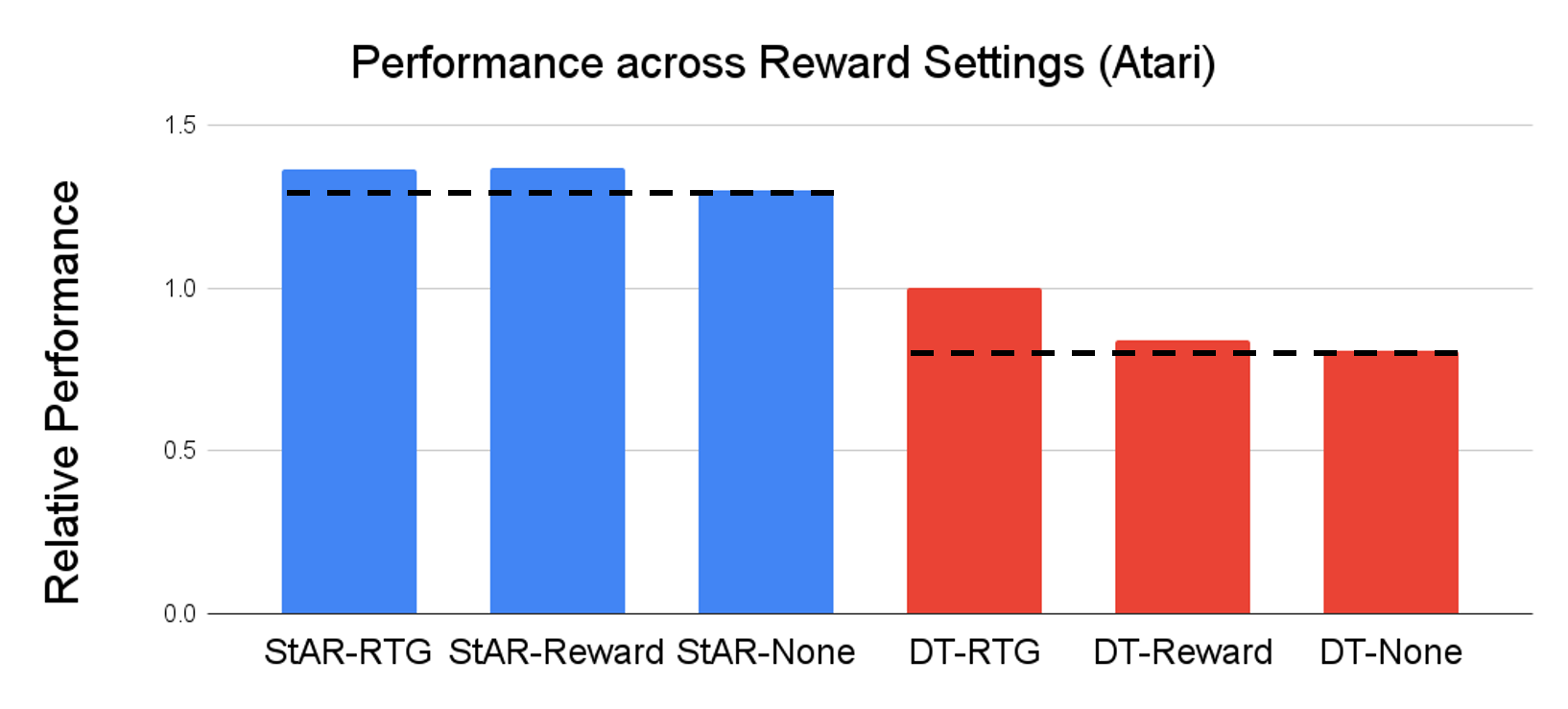}
    \caption{Performance in different reward settings: return-to-go (RTG), stepwise reward, or no reward at-all (labeled as `None') settings. 
    }
    \label{fig:reward_formulation}
\end{figure}

We investigate how different reward settings affect sequence modeling, specifically, return-to-go (RTG)~\cite{dt}, stepwise reward, and no reward at-all. Decision-Transformer~\cite{dt} originally uses RTG $\hat{R}_t$, which is defined as the sum of future step-wise rewards: $\hat{R}_t= \sum_{t'=t}^T r_{t'}$, widely used in~\cite{kaelbling1993learning,andrychowicz2017hindsight,pong2018temporal,srivastava2019upsidedown,kumar2019rewardconditioned,li2020generalized,eysenbach2020rewriting}. Stepwise reward $r_t$ is the immediate reward generated by an environment in each step, which is generally used in most RL algorithms~\cite{dqn,hessel2018rainbow}. \modelname~uses $r_t$ by default, guided by the motivation of modeling single-step transitions. No reward at-all corresponds to imitation (behavior cloning).

Results are shown in Fig.~\ref{fig:reward_formulation}.
\modelname~and DT behaves differently when reward settings are varied.
Both methods show performance gains with reward.
Also, \modelname~performs similarly regardless of RTG or stepwise reward, whereas DT relies more on RTG, and \modelname-None can still outperform DT-RTG, even without reward.
These observations tell sequence modeling can even work on state-action-only trajectories when the model has enough capacity. 
Such observation is consistent with Dreamerv2~\cite{hafner2020dreamerv2}, where no-reward setting performs as well as having reward due to the strong dynamics model.

\section{Conclusion}
In this work, we introduce \modelname, which models strong local relations explicitly (\stepname) to help improve the long-term sequence modeling (\seqname) in Visual RL. Our extensive empirical results show how the learned \repnames~help our model to outperform the baseline.
We further demonstrate that our method successfully models trajectories, with an emphasis on long sequences.

\paragraph{Acknowledgements.}
We thank members in Robotics Lab at Stony Brook for valuable discussions.
This work was supported by Institute of Information \& communications Technology Planning \& Evaluation (IITP) grant funded by the Ministry of Science and ICT (No.2018-0-00205, Development of Core Technology of Robot Task-Intelligence for Improvement of Labor Condition. This work was also supported by the National Science Foundation (IIS-2104404 and CNS-2104416).

\bibliography{stf_new}
\bibliographystyle{splncs04}

\appendix
\newpage
\section*{Appendix}
\subsection*{Detailed Results}
\begin{figure*}[htbp]
    \centering
    \includegraphics[width=\textwidth]{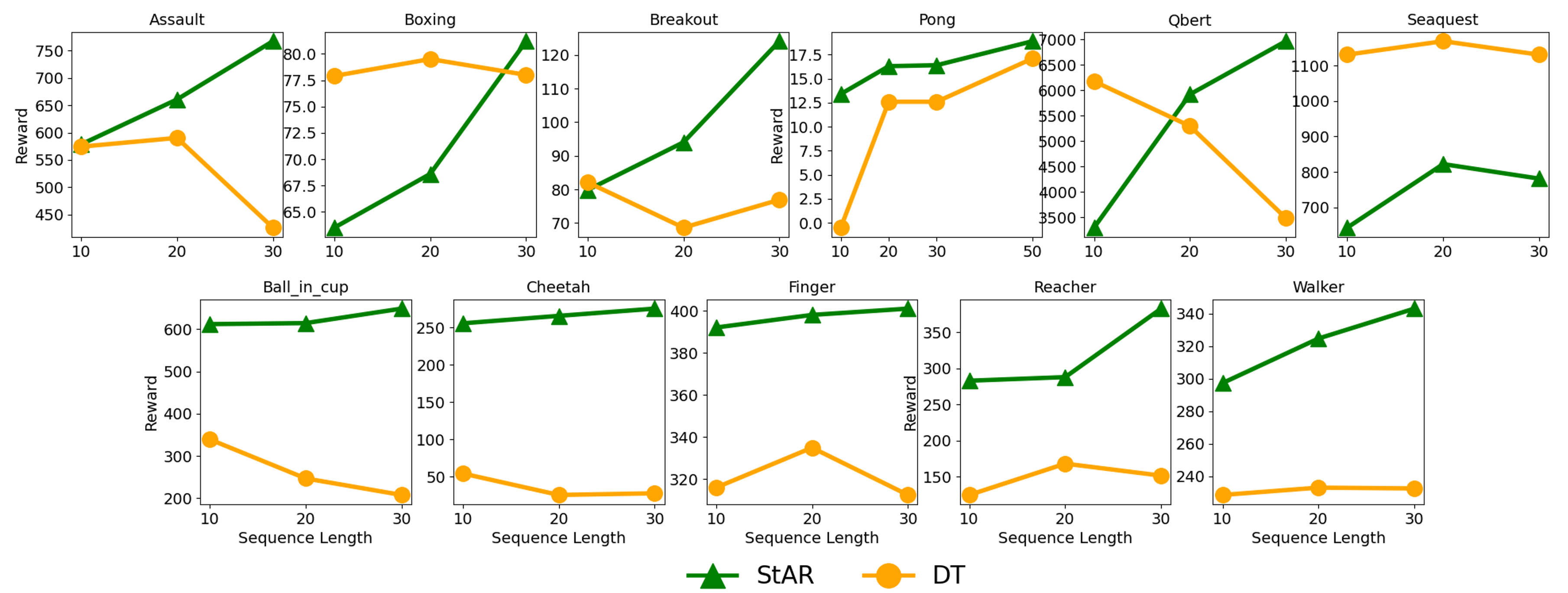}
    \caption{Performance (offline RL) under trajectory length $T\in\{10, 20, 30\}$. We also include result of $T=50$ in Pong. In most of the cases, \modelname~(green) shows a better performance than DT (yellow) when increasing the trajectory length, and surpasses that of the baseline, validating that our method can effectively model long sequences. }
    \label{fig:seqlength}
\end{figure*}

In the following Table~\ref{tab:mainres}, we give full evaluation results in exact episode returns. In More on the next page: in Fig.~\ref{fig:seqlength}, we show how each environment changes when scaling from short to relative longer sequences. Table~\ref{tab:grouping} gives per-task ablation results of embedding methods. Table~\ref{tab:connection} gives per-task ablation results of our Transformer connections.
\begin{table*}[htbp]
    \centering
    \caption{Evaluation of episodic returns ($\uparrow$) in our proposed \modelname~(StAR) and the baseline Decision-Transformer (DT)~\cite{dt} in Atari and DMC. We also compare with non-Transformer offline-RL methods: CQL~\cite{cql}, QR-DQN~\cite{qrdqn}, REM~\cite{rem}, and BEAR~\cite{bear}, in their applicable tasks (due to action space), and a behavior cloning baseline using ViT as the visual encoder (BC-ViT). The highest scores in each setting and environment are highlighted in \textbf{bold}. We use T-test to show the significance ($p<0.05$) of improvement over the baseline. We also present the max. reward of the training datasets for reference.}
    \setlength{\tabcolsep}{2pt}
    \resizebox{\linewidth}{!}{
        \begin{tabular}{llrrrrrrrrrrrrr}
        \toprule[1.3pt]
         \multirow{2}{*}{\textbf{Setting}} & \multirow{2}{*}{\textbf{Method}} & \multicolumn{7}{c}{\textbf{Atari }} & \multicolumn{6}{c}{\textbf{DMC}}\\
        \cmidrule(lr){3-9}\cmidrule(lr){10-15}
        &   & Assault   & Boxing    & Breakout  & Pong & Pong(50) & Qbert & Seaquest & & Ball\_in\_cup & Cheetah & Finger & Reacher & Walker\\
        \midrule[0.7pt]
        \multirow{7}{*}{offline RL} & DT & 462 ± 139 & 78.3 ± 4.6 & 76.9 ± 17.1 & 12.8 ± 3.2 & 17.1 ± 1.8 & 3488 ± 631 &\textbf{1131 ± 168} &  & 207.7 ± 123.2 & 27.9 ± 44.5 & 312.4 ± 94.4 & 151.4 ± 29.9 & 232.5 ± 64.8 \\
        & StAR & \textbf{761 ± 127} & \textbf{81.2 ± 3.9} & \textbf{124.1 ± 19.8} & \textbf{16.4 ± 2.1} & \textbf{18.9 ± 0.7} & 6968 ± 1698 & 781 ± 212 & & \textbf{648.4 ± 75.3} & \textbf{275.3 ± 47.9} & \textbf{401.1 ± 34.2} & \textbf{383.2 ± 59.6} & \textbf{343.1 ± 43.8}\\
        & $p$-value & 0.0005 & 0.1904 & 0.0005 & 0.0089 & 0.0461 & 0.0011 & 0.0054 & & 8.4e-8 & 5.7e-10 & 0.0170 & 4.8e-8 & 0.0004 \\
        \cmidrule(lr){2-9} \cmidrule(lr){10-15}
         & CQL & 432 & 56.2 &  55.8 & 13.5 & - & \textbf{14012} & 685 & & 176.3 & 20.3 & 264.4 & 142.6 & 78 \\
         & QR-DQN & 142 & 14.3 & 4.5 & 2.2 & - & 0.0 & 161 && - & - & - & - & -\\
         & REM & 350 & 12.7 & 2.4 & 0.0 & - & 0.0 & 282 && - & - & - & - & -\\ 
         & BEAR & - & - & - & - & - & - & - && 160.8 & 6.3 & 223.2 & 102.3 & 44 \\
          
        \cmidrule(lr){1-9} \cmidrule(lr){10-15}
        \multirow{4}{*}{Imitation} & DT & 595 ± 89 & 72.0 ± 2.6 & 54.3 ± 1.2 & 7.7 ± 2.1 & 9.7 ± 4.2 & 2099 ± 1075 & 826 ± 118 & & 319.5 ± 195.7 & 0.5 ± 0.3 & 285.2 ± 122.9 & 127.5 ± 53.5 & 230.7 ± 86.6\\
        & StAR & \textbf{788 ± 146} & \textbf{76.2 ± 3.6} & \textbf{103.1 ± 21.3} & \textbf{15.6 ± 2.6} & \textbf{17.7 ± 2.4} & \textbf{5709 ± 1002} & \textbf{939 ± 97} & & \textbf{607.7 ± 59.9} & \textbf{231.9 ± 46.2} & \textbf{400.1 ± 52.8} & \textbf{356.3 ± 76.7} & \textbf{329.9 ± 34.9} \\
        & $p$-value & 0.0014 & 0.03233 & 0.0004 & 0.0002 & 0.0029 & 2.8e-5 & 0.0430 & & 0.0011 & 7.0e-8 & 0.0185 & 8.3e-7 & 0.0058 \\
        \cmidrule(lr){2-9} \cmidrule(lr){10-15}
        & BC-ViT & 442 & 58.0 & 4.9 & -13.7 & -  & 554 & 275 & & 125.1 & 1.2 & 74.7 & 107.0 & 97.6\\
         \cmidrule(lr){1-9} \cmidrule(lr){10-15}
        \multicolumn{2}{l}{Dataset} & 153 & 98 & 92 & 21 & 21 & 600 & 290 & & 541 & 354 & 771 & 348 & 422\\
       
        \bottomrule[1.3pt]
        \end{tabular}
}
    \label{tab:mainres}
\end{table*}

\begin{table*}[hbtp]
    \centering
    \caption{Ablation results on \repnames~in \stepname~and pure state representation $h_t$ in \seqname~(offline RL, Atari). StAR-rep stands for having \repname~in the model. Emb. of $s_t$ is the method used for learning \repname. Emb. of $s_t$ is the pure state representation, which is only applicable to our method. We notice the combination of patch embeddings and CNN features works the best than other methods. Simply replace CNN features to ViT-like patch embeddings in DT will not improve the performance.}
    \resizebox{\linewidth}{!}{
    \begin{tabular}{lcccrrrrrr}
    \toprule[1.3pt]
        Method & StAR-rep. & Emb. of $s_t$ & Emb. of $h_t$ & Assault & Boxing & Breakout & Pong & Qbert & Seaquest\\
        \midrule
        StAR (P+C) & \cmark & patch & \cmark, conv & \textbf{761 ± 127} & \textbf{81.2 ± 3.9} & \textbf{124.1 ± 19.8} & \textbf{16.4 ± 2.1} & \textbf{6968 ± 1698} & 781 ± 212\\
        \cmidrule{1-10}
        P+P & \cmark & patch & \cmark, patch & 548 ± 136  & 49.4 ± 4.7 & 38.0 ± 14.1 & 13.0 ± 5.7 & 1724 ± 472 & 565 ± 191 \\
        P+\_\_ & \cmark &  patch & \xmark & 583 ± 107 & 78.8 ± 2.4 & 38.7 ± 4.7 & 13.9 ± 3.3 & 4170 ± 942 & 920 ± 130 \\
        C+C & \cmark & conv & \cmark, conv & 694 ± 31 & 52.5 ± 5.0 & 55.6 ± 12.6 & 11.0 ± 4.0 & 3505 ± 2132 & 844 ± 274\\
        C+P & \cmark & conv & \cmark, patch & 285 ± 35 & 65.0 ± 3.6 & 46.3 ± 17.9  & 14.3 ± 2.3  & 3529 ± 2545 & 568 ± 14 \\
        C+\_\_ & \cmark & conv & \xmark & 509 ± 74 & 65.3 ± 4.5 & 52.7 ± 11.5 & 10.7 ± 3.5 & 1100 ± 554 & 853 ± 71\\
        \cmidrule{1-10}
        DT with ViT. & \xmark & - & patch & 608 ± 85 & 74.3 ± 13.8 & 47.3 ± 7.4 & 2.7 ± 16.5 & 1135 ± 585 & 885 ± 93\\ 
        DT~\cite{dt} &  \xmark & - & conv & 504 ± 54 & 78.3 ± 4.6 & 70.7 ± 8.1 & 12.8 ± 3.2 & 3782 ± 695 &\textbf{1007 ± 170} \\
        \bottomrule[1.3pt]
    \end{tabular}
    }
    \label{tab:grouping}
\end{table*}

\begin{table}[htbp]
    \centering
    \setlength{\tabcolsep}{1.5pt}
    \caption{Ablation results on Transformer connectivity (offline RL). We observe that our original structure (StAR), which is a layer-wise manner, fits more than StAR Fusion and StAR Stack connections shown by higher rewards.}
    \resizebox{\linewidth}{!}{
    \begin{tabular}{lrrrrrr}
    \toprule
         Method  & Assault & Boxing & Breakout & Pong & Qbert & Seaquest\\
         \midrule
         StAR & 761 ± 127 & \textbf{81.2 ± 3.9} & \textbf{124.1 ± 19.8} & \textbf{16.4 ± 2.1} & \textbf{6968 ± 1698} & \textbf{781 ± 212}\\
         StAR Fusion & 756 ± 116  & 69.4 ± 2.8 & 29.2 ± 15.9 & 8.7 ± 5.1 & 4053 ± 1239 & 608 ± 174 \\
         StAR Stack& \textbf{939 ± 157} & 64.9 ± 4.9  & 30.9 ± 5.5 & 13.7 ± 2.5 & 575 ± 124 & 361 ± 261 \\
     \bottomrule
    \end{tabular}
    }
    \label{tab:connection}
\end{table}

\newpage
\subsection*{More Visualizations}
\begin{figure}[h]
    \vspace{-10pt}
    \centering
    \includegraphics[width=0.56\linewidth]{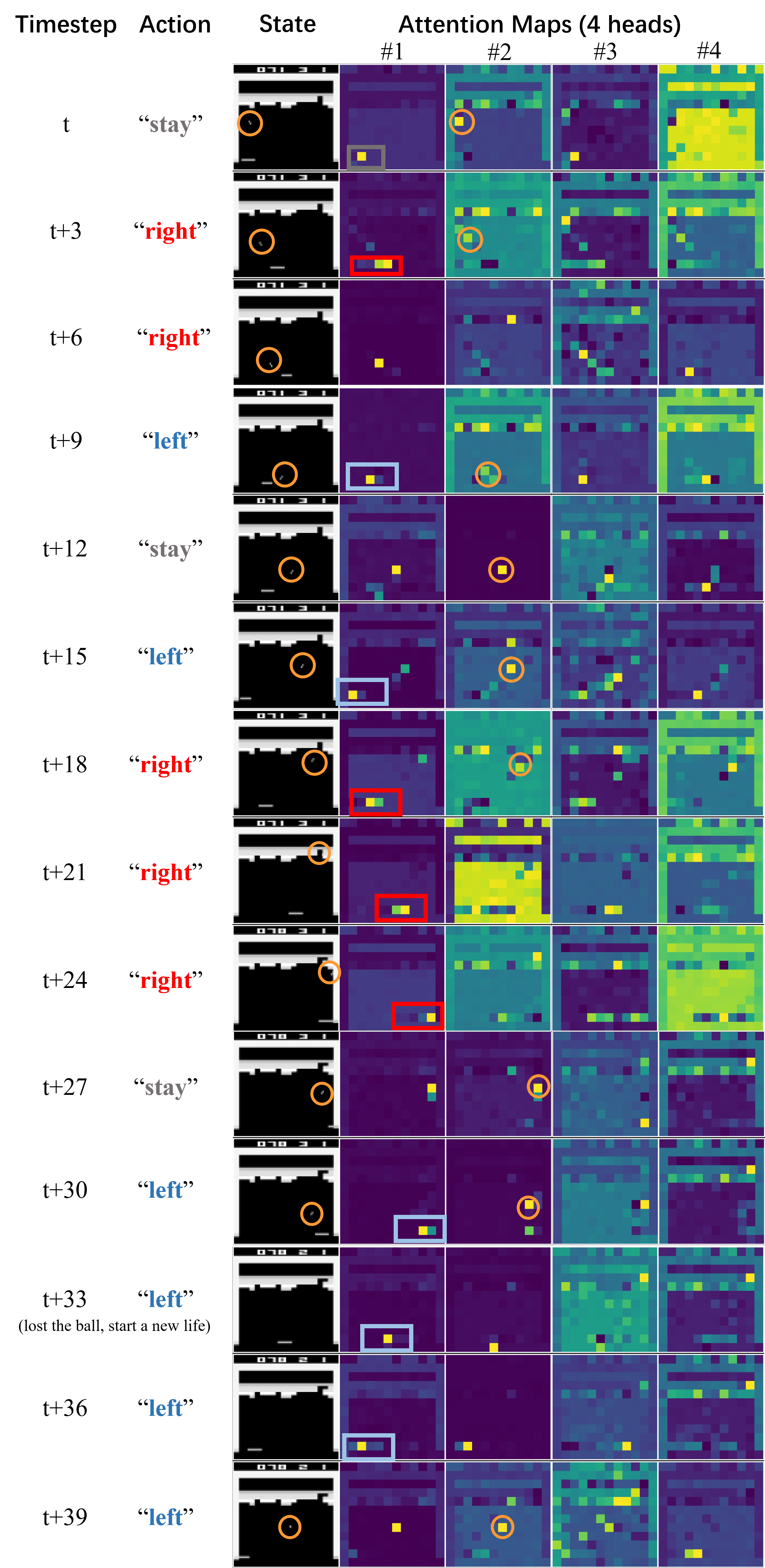}
    \caption{More attention maps visualization in our \stepname.  }
    \label{fig:vis_more}
\end{figure}

We visualize a segment of real trajectory in Breakout in Fig.~\ref{fig:vis_more}, with annotated actions and highlighted labels for easier understanding. In general, we observe higher attention scores at the areas where the paddle and the ball locate in different attention heads. We also find relatively consistent semantic meanings in attention heads \#1 and \#2, with focus on pad and ball, respectively. 

\newpage
\subsection*{Hyper-parameters}
The complete list of hyper-parameters are given in Table~\ref{tab:hp} and \ref{tab:hp2}, for Atari and DMC respectively.
We keep most of the hyper-parameters similar to those provided by Decision-Transformer~\cite{dt} for a fair comparison, including the number of Transformer layers, MSA heads and embedding dimensions in our \seqname, learning rate and optimizer configurations. Since DT does not conduct experiments in DMC environment with visual input, we tune DT and set learning rate to be $1\times 10^{-3}$ in DMC. For the frame skipping in DMC, we use the setting from~\cite{sacae}, for both DT and our method.
\begin{table}[htbp]
    \centering
    \caption{Our hyper-parameter settings in Atari. Underlined parameters are unique/different from DT~\cite{dt}}
    \resizebox{\linewidth}{!}{
    \begin{tabular}{ll}
    \toprule
        Hyper-parameter & Value \\
        \midrule
        Input sequence length ($T$) & 10, 20, 30\\
        Input image size & $84\times84$, gray \\
        Frame stack & 4 \\
        Frame skip & 2 \\
        Layers & 6 \\
        MSA heads (\seqname) & 8 \\
        Embedding dimension (\seqname) & 192 \\
        \underline{Image patch size} & 7 \\
        \underline{MSA heads (\stepname)} & 4 \\
        \underline{Embedding dimension (\stepname)} & 64 \\
        Nonlinearity & GeLU for self-attention; ReLU for convolution \\
        Dropout & 0.1 \\
        Learning rate & $6\times10^{-4}$ \\
        Adam betas & (0.9, 0.95) \\
        Grad norm clip & 1.0 \\
        Weight decay & 0.1 \\
        Learning rate decay &  Linear warmup and cosine decay (see~\cite{dt}) \\
        Warmup tokens & $512 \times 20$ \\
        Final tokens & $2 \times 500000 \times T$\\
        \bottomrule
    \end{tabular}
    }
    \label{tab:hp}
\end{table}

\begin{table}[htbp]
    \centering
    \caption{Our hyper-parameter settings in DMC. Underlined parameters are unique/different from DT~\cite{dt}}
    \resizebox{\linewidth}{!}{
    \begin{tabular}{ll}
    \toprule
        Hyper-parameter & Value \\
        \midrule
        Input sequence length ($T$) & 10, 20, 30\\
        Input image size & $84\times84$, gray \\
        Frame stack & 3 \\
        Frame skip & 4 for Cheetah and Reacher, 2 for Walker \\
        Layers & 6 \\
        MSA heads (\seqname) & 8 \\
        Embedding dimension (\seqname) & 192 \\
        \underline{Image patch size} & 12 \\
        \underline{MSA heads (\stepname)} & 4 \\
        \underline{Embedding dimension (\stepname)} & 64 \\
        Nonlinearity & GeLU for self-attention; ReLU for convolution \\
        Dropout & 0.1 \\
        Learning rate & $1\times10^{-3}$ \\
        Adam betas & (0.9, 0.95) \\
        Grad norm clip & 1.0 \\
        Weight decay & 0.1 \\
        Learning rate decay &  Linear warmup and cosine decay (see~\cite{dt}) \\
        Warmup tokens & $512 \times 20$ \\
        Final tokens & $2 \times 100000 \times T$\\
        \bottomrule
    \end{tabular}
    }
    \label{tab:hp2}
\end{table}

\end{document}